\documentclass[letterpaper]{article}

\usepackage{aaai}
\usepackage{times}
\usepackage{helvet}
\usepackage{courier}
\usepackage{xspace}
\usepackage{color}
\usepackage{hyperref}
\hypersetup{bookmarksopen,bookmarksnumbered,
pdfpagemode=UseOutlines,
colorlinks=true,
linkcolor=blue,
anchorcolor=blue,
citecolor=blue,
filecolor=blue,
menucolor=blue,
urlcolor=blue
}
\usepackage{amsmath,amssymb,amsfonts,amsfonts}
\usepackage{algorithm}
\usepackage[noend]{algorithmic}
\usepackage{url}
\usepackage{graphicx,subfigure}

\usepackage{float}
\usepackage{bm}

\usepackage{multirow}

\newenvironment{proof}{\paragraph{Proof:}}{\hfill$\square$}

\title{\LARGE \bf
Efficient motion planning for problems lacking optimal substructure
}
\author{
Oren Salzman \and 
Brian Hou \and
Siddhartha Srinivasa \\
The Robotics Institute 
Carnegie Mellon University Pittsburgh, PA
\thanks{
This work was (partially) funded by the National Science Foundation IIS (\#1409003), Toyota Motor Engineering \& Manufacturing (TEMA), and the Office of Naval Research.
}
\thanks{
\texttt{\{osalzman, bhou1,ss5\}@andrew.cmu.edu}.
}}

\newcommand{\calX}{\ensuremath{\mathcal{X}}\xspace}

\newcommand{\calQ}{\ensuremath{\mathcal{Q}}\xspace}
\newcommand{\calT}{\ensuremath{\mathcal{T}}\xspace}

\newcommand{\R}{\mathbb{R}}

\newcommand{\Cfree}{\ensuremath{\calX_{\rm free}}\xspace}
\newcommand{\Cforb}{\ensuremath{\calX_{\rm obs}}\xspace}
\newcommand{\Crisk}{\ensuremath{\calX_{\rm risk}}\xspace}
\newcommand{\Csafe}{\ensuremath{\calX_{\rm safe}}\xspace}

\newtheorem{thm}{Theorem}
\newtheorem{lem}{Lemma}
\newtheorem{observation}[thm]{Observation}

\newcommand{\ignore}[1]{}

\newcommand{\arxiv}[2]{#1}

\newboolean{ShowAuthors}
\setboolean{ShowAuthors}{false}
\def\blind#1{
\ifthenelse{\boolean {ShowAuthors}}{#1}{}
}

\newcommand{\captionstyle}{\sf \footnotesize }

\begin{document}

\maketitle
\thispagestyle{empty}
\pagestyle{empty}

\begin{abstract}
We consider the motion-planning problem of planning a collision-free path of a robot in the presence of risk zones.
The robot is allowed to travel in these zones but is penalized in a super-linear fashion for consecutive accumulative time spent there.
We suggest a natural cost function that balances path length and risk-exposure time. 
Specifically, we consider the discrete setting where we are given a graph, or a roadmap, and we wish to compute the minimal-cost path under this cost function. Interestingly, paths defined using our cost function do not have an optimal substructure.
Namely, subpaths of an optimal path are not necessarily optimal. 
Thus, the Bellman condition is not satisfied and standard graph-search algorithms such as Dijkstra cannot be used.
We present a path-finding algorithm, which can be seen as a natural generalization of Dijkstra's algorithm.
Our algorithm runs in $O\left((n_B\cdot n) \log( n_B\cdot n) + n_B\cdot m\right)$ time, where~$n$ and $m$ are the number of vertices and edges of the graph, respectively,  and $n_B$ is the number of intersections between edges and the boundary of the risk zone.
We present simulations on robotic platforms demonstrating both the natural paths produced by our cost function and the computational efficiency of our algorithm.
\end{abstract}

\section{Introduction}
\label{sec:introduction}

In this paper, we explore motion-planning problems where an agent has to compute the least-cost path to navigate through \emph{risk zones} while avoiding obstacles.
Travelling these regions incurs a penalty which is \emph{super-linear} in the traversal time.
We call the class of problems 
\emph{Risk Aware Motion Planning (RAMP)}
and define a natural cost function which simultaneously optimizes for paths that are both short and reduce consecutive exposure time in the risk zone.

We are motivated by real-world problems involving \emph{risk}, where 
 continuous exposure is much worse than intermittent exposure.
Examples include pursuit-evasion
where sneaking in and out of cover is the preferred strategy,
and visibility planning where the agent must ensure that
an observer or operator is minimally occluded. 

Interestingly this setting 
(where  continuous exposure is much worse than intermittent exposure) is not limited to planning:
consider a large number of processes running in a shared-memory environment. Here, we would like to \emph{quantify} the cost of running each process.
When a process writes to the shared memory, it is required to use a mutex mechanism, possibly blocking other processes. 
Clearly, as the blocking-time increases, the probability that other processes have to stay idle, increases.
Thus, as in the previous examples, the cost of running the process is its duration with a super-linear penalty proportional to accumulative consecutive times where the process writes to the shared memory.

\begin{figure}[t!]
  \centering
    \includegraphics[height = 2.8cm ]{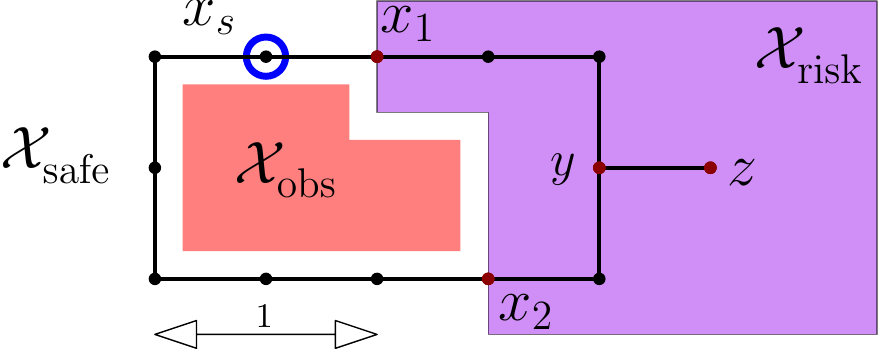}
  \caption{
  \captionstyle
  Example for which optimal paths
 computed using our cost function (Eq.~\ref{eq:cost})
  do not have an optimal substructure.
  Distance between two consecutive grid points is 0.5.
  The minimal-cost path 
  from $x_s$ to~$y$ passes through $x_1$ and has a 
  cost of $0.5 + (e^{1.5} - 1) \approx 3.98$ 
  while the path 
  passing through~$x_2$ has a 
  cost of $3 + (e^{1} - 1) \approx 4.78$.
  In contrast, the minimal-cost path to $z$ 
  does path through  $x_2$ and has a 
  cost of $3 + (e^{1.5} - 1) \approx 6.48$ 
  while the path 
  passing through   $x_1$  has a 
  cost of $0.5 + (e^{2} - 1) \approx 6.88$.
    }
    \label{fig:substructure}
\vspace{-5mm}
\end{figure}

Although practically useful, our cost function for RAMP suffers from one
fundamental algorithmic challenge: 
\emph{optimal plans do not posses optimal substructure}.

We explain with an example and an analogy.
Consider Fig.~\ref{fig:substructure}, where an agent, traversing a graph, starts at $x_s$ and must reach $z$ while avoiding $\Cforb$.
The cost of traversing through
$\Csafe$ is linear in the distance travelled 
while the cost of traversing $\Crisk$ is super-linear in the distance travelled
(formalized in Sec.~\ref{sec:problem_formulation}).

Now consider two snails (an homage to Pohl~\cite{pohl1969bi}) taking
two different paths racing to $z$: snail $A$ passes through~$x_1$ and $y$, 
and snail $B$ through $x_2$ and $y$.
If there were no risk zones, and the snails move at constant speed, the optimal substructure implies that when~$B$ reaches~$y$ and notices that $A$ has passed through it already
(snails leave personalized slime trails)
it has no hope of catching up and should give up. In other words,
the first snail that reaches $z$ is \emph{also} the first snail that reaches
every intermediate point along the optimal path to $z$.
Optimal substructure is critical for search algorithms~\cite{D59} to efficiently track all ``promising'' snails (Sec.~\ref{sec:prelim}).

Unfortunately, our cost function for RAMP does \emph{not} posses optimal substructure.
Imagine that snails passing through a risk region accumulate muck on their foot.
The more time a snail spends in a risk region, the more muck it accumulates, and the \emph{slower} it gets. 
Now, although $B$ reaches $y$ later than $A$,
it has less muck on it, having spent less time in the risk zone than $A$, 
and actually \emph{catches up and overtakes} $A$ to reach $z$ first.

We define a \emph{boundary point} to be at the 
boundary of a risk-free and a risk region 
(like $x_1$ and $x_2$ in our example)
and let~$n_B$  denote the 
number of 
boundary points in our graph.

We can now address RAMP via two key insights:
\begin{enumerate}
  \item Any optimal path from a risk-free start to a risk-free goal that passes through a risk region can be decomposed into 
  (i)~optimally reaching a boundary point, 
  (ii)~optimally traversing the risk region, and 
  (iii)~optimally  reaching the goal from the exit (of the risk region). As a consequence, if we store $O(n_B^2)$ pairwise optimal paths (one for each pair of boundary points), we could create an augmented graph which can be used to directly solve RAMP. 
  \item All snails that pass through one specific boundary point satisfy the optimal substructure condition. In other words, the only way a snail can overtake another is if it enters the risk region via a different boundary point. As a consequence, we \emph{only} need to track $O(n_B)$ extra snails.
\end{enumerate}

Fuelled by these insights, we propose two new fundamentally different algorithms (Sec.~\ref{sec:alg})
that solve RAMP efficiently by augmenting different data structures used by Dijkstra's algorithm:

Our \emph{precomputation-based algorithm}  creates an augmented graph
with boundary vertices and their precomputed optimal paths added to the original
graph, incurring a time complexity of $O\left(
n_B n \cdot \log n
+
n_B^2 \cdot \log n
+
n_B m 
\right)$. Here~$n$ and $m$ are the number of vertices and edges on the graph, respectively.

Our \emph{incremental algorithm} is a strict generalization of Dijkstra's algorithm to account for risk regions. It maintains an augmented priority queue to account for the $O(n_B)$ boundary points, incurring a time
complexity of $O\left(n_B n \cdot \log n +  n_B m\right)$. 

Interestingly the (asymptotic) running time of both algorithms is identical 
unless the $n_B^2 \log n$ component of the precomputation 
dominates the running time.
This happens when $n_B \log n= \omega(m)$.
Intuitively, the precomputation algorithm slows down if 
there are many boundary \emph{vertices}, whereas
the incremental algorithm slows down if there are many risk-zone \emph{edges}.

We evaluate our algorithms in Sec.~\ref{sec:eval}.
Our experiments show that our cost function naturally balances path length and risk-exposure times. Furthermore we demonstrate the advantage of our incremental algorithm over the precomputation-based approach.
Finally, we discuss future work in Sec.~\ref{sec:conclusions}.

\section{Related work}
\label{sec:related_work}
The RAMP problem lies on the intersection between several disciplines which we will briefly review.
In its general form, our problem can be seen as an instance of the \emph{motion-planning problem}~\cite{L06,CBHKKLT05}.
Indeed, in this work we follow the \emph{sampling-based paradigm} (Sec.~\ref{subsec:mp}).
This calls for computing a discrete graph which is then traversed by a \emph{path-finding} algorithm (Sec.~\ref{subsec:pp}).

Computing minimal-risk paths in high-dimensional spaces is related to motion-planning  under \emph{risk constraints} (Sec.~\ref{subsec:risk}).
When the search domain is two dimensional, alternative, more efficient approaches exist  (Sec.~\ref{subsec:low}).

\subsection{Sampling-based motion planning}
\label{subsec:mp}
The basic motion-planning problem calls for moving a robot~$R$ in a workspace cluttered with static obstacles
from a start position to a target one while avoiding obstacles and minimizing some cost function.
Typically, $R$ is abstracted as a point in a \emph{configuration space}~\calX which is subdivided into free and forbidden regions~\cite{L83}.
The problem then reduces to computing a minimal-cost collision-free path for a point in~\calX.

For high-dimensional problems, even computing a path, let alone an optimal one, becomes computationally hard~\cite{R79}.
Thus, a common approach is to approximate $\calX$ using a graph, or a roadmap,~$G$. 
One such example is the Probabilistic Roadmap Planner or PRM~\cite{KSLO96}. Here vertices of~$G$ are points sampled in $\calX$ and two ``close-by'' vertices are connected by an edge if the straight line connecting the two does not intersect obstacles in $\calX$. 
A query is then answered by running a shortest-path algorithm on $G$.
Under certain assumptions, the cost of solutions obtained by this algorithm converge to the cost of the optimal solution as the number of samples grow~\cite{KF11,SSH16c}.

\subsection{Path planning}
\label{subsec:pp}
Planning a shortest path on a given graph is a well studied problem.
Given a graph with a cost function on its edges, the shortest-path problem asks for finding a path of minimum cost between two given vertices. 
When the cost function has an optimal substructure, efficient algorithms such as Dijkstra~\cite{D59}, A*~\cite{HNR68}  and their many variants can be used.

In certain applications, including our setting, this is not the case. For example in~\cite{TZ04} every edge is associated with two attributes, say cost and resource,
and there is a  non-linear objective function which is convex and non-decreasing

Our problem is also closely related to multi-objective path planning.
Here, we are given a set of cost functions and we are interested in finding a set of paths that captures the trade-off (the so-called Pareto curve) among the several objectives.
In general, the number of efficient solutions may be exponential in the problem size~\cite{RG00}.
However,  a fully polynomial-time approximation scheme (FPTAS) can be found, even for the case where the cost functions are non additive~\cite{TZ06}.
For recent results on multi-objective and multi-constrained non-additive shortest path problems,
\arxiv{see e.g.~\cite{RP11,CN13}}{see e.g.~\cite{RP11}}
and references within.

\subsection{Planning under risk constraints}
\label{subsec:risk}
Planning under risk constraints has been studied in several motion-planning settings.
A common  approach to formulate this problem is to assign 
some risk values to regions.
Paths are considered by the planning algorithm only if the risk obtained in different regions are below some predefined thresholds~\cite{OWB13,OPKB15,CGJP15}.

Planning under risk constraints was also considered for the specific problem of Autonomous Underwater Vehicle (AUV).
Here the cost function used was the sum of risk values at waypoints along a given path and the domain was two-dimensional~\cite{PBHS13}.

Planning under risk constraints is closely related to the problem of planning under uncertainty.
A common approach is to minimize the estimated collision probability of a path~\cite{LA14,STvA2016}.
Additional constraints are often added such as requiring  some notion of smoothness~\cite{MS14}.

\subsection{Planning in low dimensions}
 \label{subsec:low}
When planning occurs in low dimensions, efficient algorithms exist (see, e.g.,~\cite{HSS16} for a survey). 
We briefly mention several variants which are related to the problem we consider.
 
One such example is computing minimal-cost paths for weighted planar regions~\cite{MP91}.
Here, a planar space is subdivided into different regions, where each region is assigned a positive weight.
The length of a path is defined as the weighted sum of (Euclidean) lengths of the subpaths within each
region.
The problem is conjectured to be computationally hard, 
thus the focus of the community has been on approximate algorithms.
For an overview of recent results, as well as a generalization of the problem, see~\cite{SR07,JTG14}.

Another example is computing paths  which are  simultaneously short and stay away from the obstacles~\cite{WBH08}. 
It is not clear if this problem is NP-Hard, however an FPTAS for this problem is known~\cite{AFS16}.

\section{Problem formulation}
\label{sec:problem_formulation}
\begin{figure*}[t]%
  \centering%
  \subfigure[]
  {
  \includegraphics[width=0.235\textwidth]{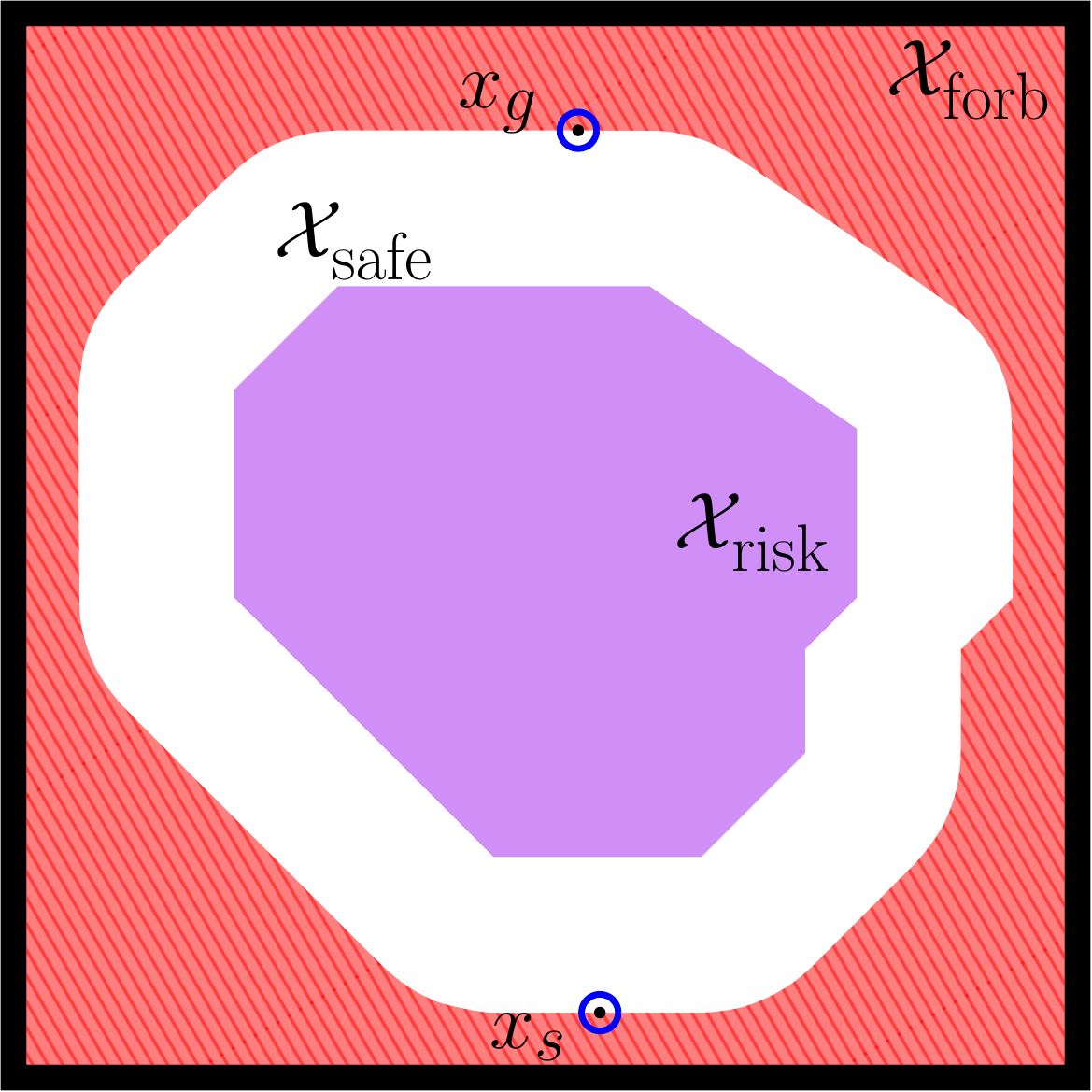}
  \label{fig:workspace}
  }
  \subfigure[]
  {
  \label{fig:roadmap}
  \includegraphics[width=0.235\textwidth]{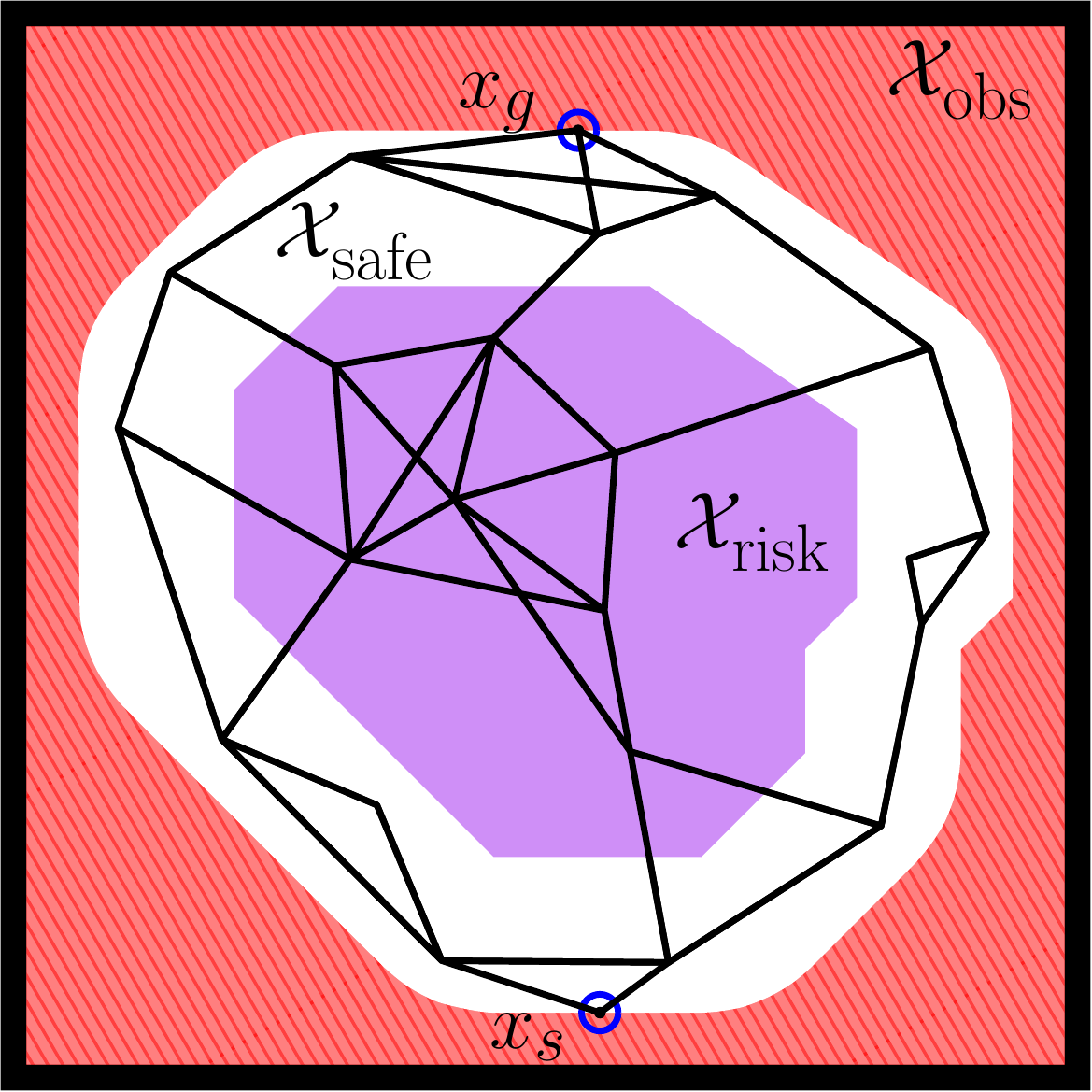}
  }
  \subfigure[]
  {
  \label{fig:refined}
  \includegraphics[width=0.235\textwidth]{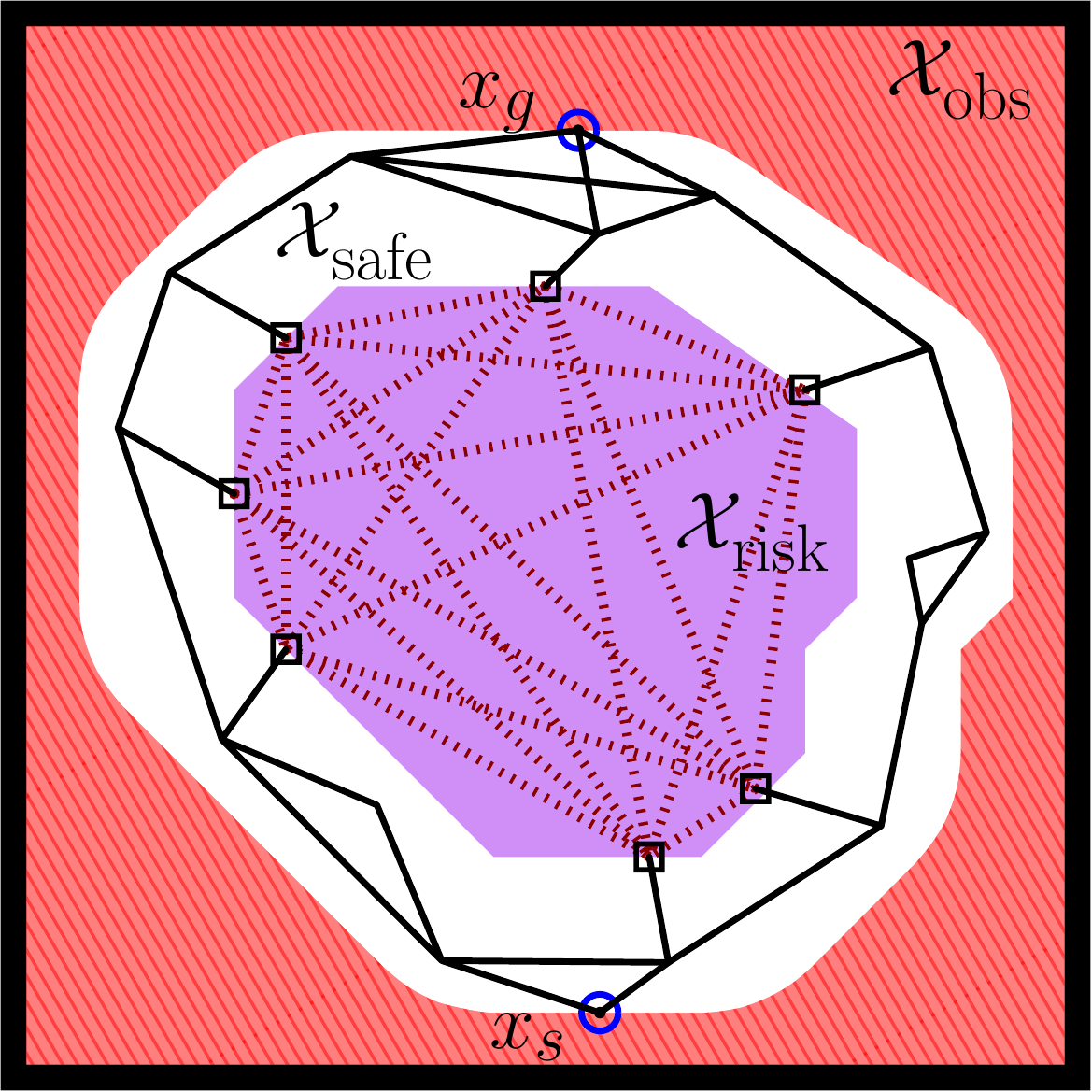}
  }  
  \subfigure[]
  {
  \label{fig:path}
  \includegraphics[width=0.235\textwidth]{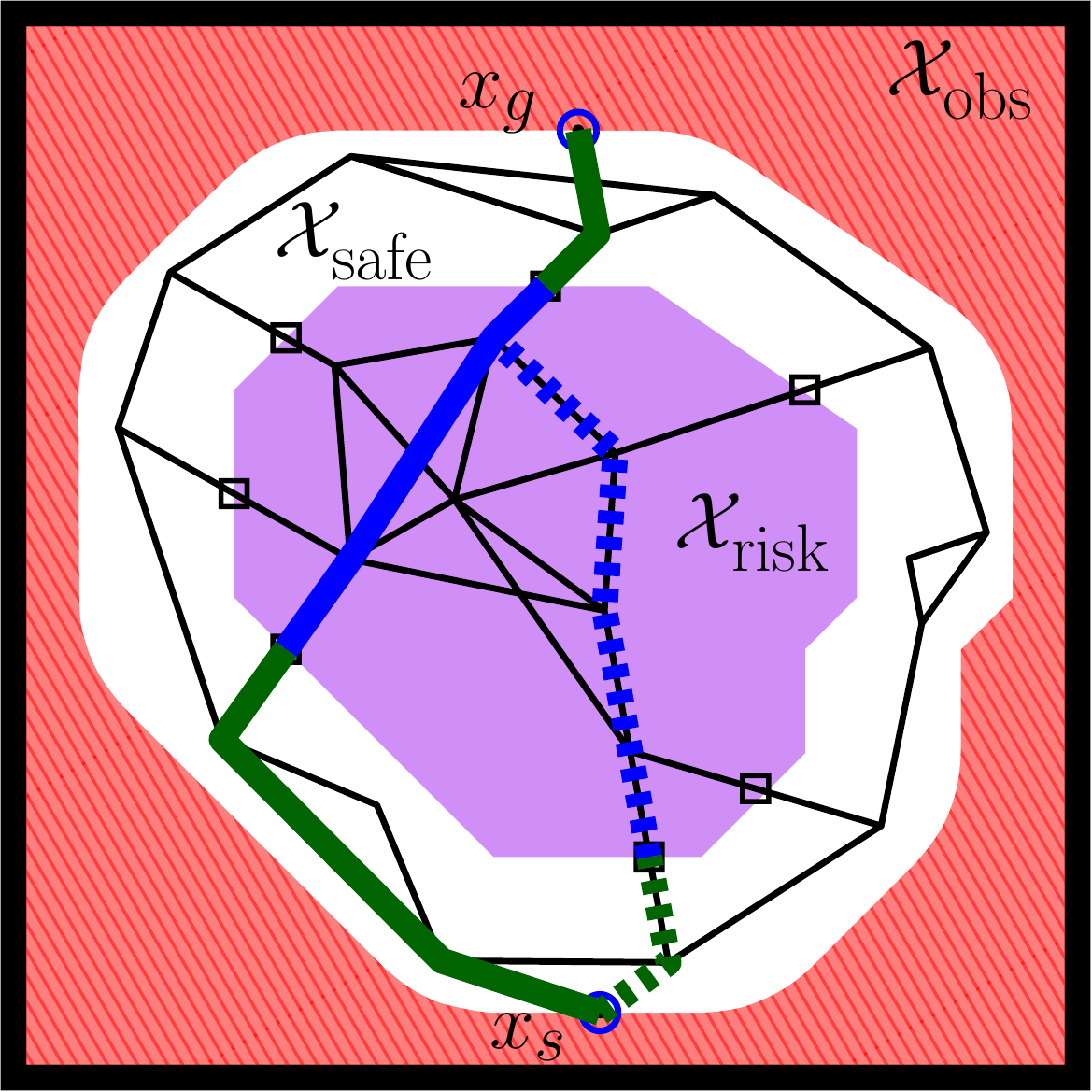}
  }
  \caption{%
    \captionstyle
         \subref{fig:workspace}~A two-dimensional space~$\calX$ consisting of obstacles (red polygons) and a risk region (purple region), defined as all points which are farther away from the obstacles than a predefined distance  (see Sec.~\ref{sec:eval} for a motivation regarding this scenario).
       \subref{fig:roadmap}~Probabilistic roadmap~$G$ sampled in~$\calX$.
       \subref{fig:refined}~The roadmap~$G'$ built by taking~$G$, adding all border points (hollow squares) as vertices and replacing all existing edges in $\Crisk$ with an edge between every pair of border points (red polylines) representing the cost of travelling between the two in $G \cap \Crisk$. 
       \subref{fig:path}~Minimal-cost path (green and blue for edges in \Csafe and \Crisk, respectively). Notice that this is \emph{not} the shortest path, which has high exposure to~\Crisk (depicted in dashed green and dashed blue).
  }%
  \label{fig:filmstrip}%
  \vspace{-2.5mm}
\end{figure*}

\begin{figure}[tb]
  \centering
  	\includegraphics[height = 4.5cm ]{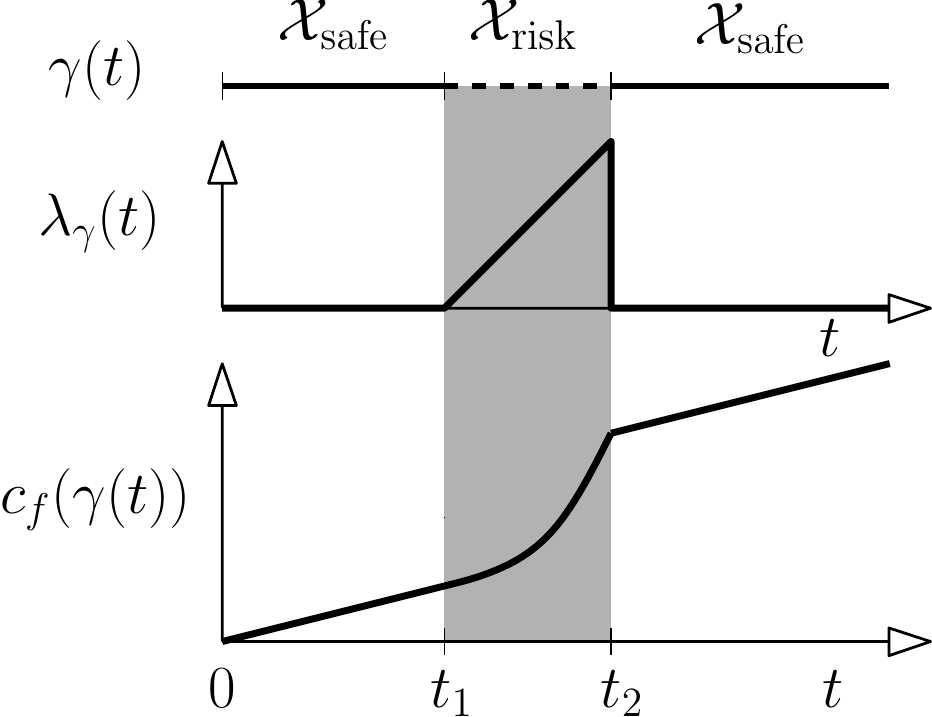}
  \caption{
    \captionstyle
  	Relation between a trajectory~$\gamma(t)$ (top), recent exposure time~$\lambda_\gamma(t)$ (middle) and cost~$c_f(\gamma(t))$ (bottom) as a function of time.
  	In $t \in [0,t_1]$, $\gamma$ stays in~\Csafe, hence~$\lambda_\gamma(t)=0$ and the cost grows linearly with time.
  	At~$t=t_1$, $\gamma$ enters~\Crisk,~$\lambda_\gamma(t)$ grows linearly and the cost grows super-linearly. 
  	At $t=t_2$, $\gamma$ leaves~\Crisk,~$\lambda_\gamma(t)=0$ and the cost returns to growing linearly.   	
  	}
   	\label{fig:cost}
	\vspace{-5.5mm}
\end{figure}

Let~$\calX$ denote the $d$-dimensional configuration space, $\Cfree$ the collision-free portion of $\calX$ and $\Cforb = \calX \setminus \Cfree$. 
Let 
$\Crisk \subset \Cfree$
and
$\Csafe = \Cfree \setminus \Crisk$
denote the risk and risk-free zones, respectively.
We assume that 
$\Crisk$
and
$\Cfree$ are open and closed sets, respectively. 
See Fig.~\ref{fig:workspace}.

A trajectory $\gamma: [0,T_\gamma] \rightarrow \Cfree$ is a continuous  mapping between time and configurations. 
The image of a trajectory is called a path.
By a slight abuse of notation we 
refer to $\gamma[t',t'']$
as the sub-path connecting $\gamma(t')$ and $\gamma(t'')$
for
$0 \leq t' \leq t'' \leq T_\gamma$.
Finally, we assume that 
both endpoints
of the  path lie in the risk-free zone. 
Namely, $\gamma(0), \gamma(T_\gamma) \in \Csafe$.

Given a trajectory $\gamma$, 
and some time $t \in [0, T_\gamma]$,
let $t' \leq t$ be the latest time such that $\gamma(t') \in \Csafe$.
Notice that if $\gamma(t) \in \Csafe$ then $t'=t$.
We define the \emph{current exposure time} of $\gamma$ at $t$ as $\lambda_\gamma(t) = t - t'$.
Namely, if $\gamma(t) \in \Crisk$ then $\lambda_\gamma(t)$ is the time passed since $\gamma$ last entered \Crisk.
If $\gamma(t) \in \Csafe$ then $\lambda_\gamma(t)=0$.

We are now ready to define our cost function.
Let 
$\gamma$ be a trajectory and 
$f(x)$ any function such that 
$f(x) = \omega(x)$ and~$f(0) = 1$.
The cost of $\gamma$, denoted by~$c_f (\gamma)$ is defined as 
\begin{equation}
\label{eq:cost}
 c_f (\gamma)
 =
 \int_{t \in [0,T_\gamma]}
 	f(\lambda_{\gamma}(t)) |\dot{\gamma(t)}| dt.
\end{equation}
Eq.~\ref{eq:cost} penalizes continuous exposure to risk in a super-linear fashion (hence the requirement that $f(x) = \omega(x)$). 
As~$f(0) = 1$, the cost of traversing the risk-free zone is simply path length.
See Fig.~\ref{fig:cost} for a conceptual visualization of the current exposure time and our cost function.

Equipped with 
our cost function we can formally state the 
risk-aware motion-planning problem:

\vspace{2mm}

\noindent
\textbf{P1} \emph{Risk-aware motion-planning problem (RAMP)}
Given the tuple 
$(\Csafe, \Crisk, \Cforb, x_{\text{s}}, x_{\text{g}}, f)$,
where $x_{\text{s}},  x_{\text{g}} \in \Csafe$ are 
start and target configurations,
compute
$\underset{\gamma \in \Gamma}{\arg\min} \ c_f (\gamma)$
with~$\Gamma$ the set of all collision-free trajectories connecting $x_{\text{s}}$ and
$x_{\text{g}}$

\vspace{2mm}

As mentioned, RAMP is a generalization of the motion-planning problem (where there are no risk zones) 
which is known to be PSPACE-Hard~\cite{R79}. 
Thus, in this paper we concentrate on the discrete version of the problem:

\vspace{2mm}

\noindent
\textbf{P2} \emph{discrete Risk-aware motion-planning problem (dRAMP)}
Given the tuple 
$(\Csafe, \Crisk, \Cforb, x_{\text{s}}, x_{\text{g}}, G , f)$,
where $G = (V,E)$ is a roadmap  embedded in the C-space such that $x_{\text{s}}, x_{\text{g}} \in V$,
compute 
$\underset{\gamma \in \Gamma_G}{\arg\min} \ c_f (\gamma)$
with $\Gamma_G$ the set of all collision-free paths\footnote{Note that we use paths to define curves in $\calX$ and as sequence of edges in a graph. These definitions coincide as the graph is embedded in $\calX$.} in $G$ connecting $x_{\text{s}}$ and $x_{\text{g}}$.
See Fig.~\ref{fig:roadmap}.

To simplify the discussion,
in the rest of this paper we assume that the robot is moving in constant speed and we use $f(x) = e^x$.
Thus, we can re-write Eq.~\ref{eq:cost} as
\begin{equation}
\label{eq:e-cost}
 c (\gamma)
 =
 \int_{t \in [0,T_\gamma]}
 	e^{\lambda_{\gamma}(t)} dt.
\end{equation}
Using the assumption that the robot is moving in constant speed, we will use the terms 
duration of a trajectory and path length interchangeably
(here we measure path length as the Euclidean distance).
Further exploiting this assumption and  by a slight abuse of notation we will also use Eq.~\ref{eq:e-cost} to define the cost of a path (and not of a trajectory).

\section{Preliminaries}
\label{sec:prelim}
In this section we review Dijkstra's path-finding algorithm (using a min-priority queue) which relies on the cost function to have an optimal substructure.
We then continue to discuss our cost function---why it does not have an optimal substructure and what properties it does have.

\subsection{Dijkstra's shortest-path algorithm}
Dijkstra's algorithm computes the minimal-cost path between a given start vertex $x_s$ and all other vertices in a graph.
Returning to our snails, the algorithm can be intuitively described as follows: 
a snail starts at $x_s$ moving at constant speed. Every time it reaches a vertex, it splits into multiple snails, one for each outgoing edge. If a snail reaches a vertex which was already reached by another snail (identified by the existing trail of slime), it retracts into its shell and stops moving. Clearly, the distance travelled by the first snail to reach any vertex~$u$ is the minimal distance to reach~$u$.

The problem with the aforementioned process is that it is continuous. Dijkstra's algorithm uses discrete times where each time-step represents the event that a snail reaches a specific vertex.
The key difference is that here only one snail moves at a time and his movement spans the entire length of an edge.
Processing an event is similar to the continuous version: if a snail reaches a vertex which was already reached by another snail, it retracts into it's shell and stops its progress.
Otherwise, it splits into multiple snails, one for each outgoing edge, and the time the snail is intended to reach the edge's endpoint is computed. This is registered as a new event.

The implementation of the aforementioned process
is via a min-cost priority queue $\calQ$. Each entry $\tau = (u,c,p)$ in the queue represents the time, or cost,~$c$ that a snail is to reach the vertex $u$ through the parent $p$. 
We emphasize that for each vertex $u$, only the \emph{current best} path (or snail) is maintained, together with its cost. 
Initially, this value is only known for $x_s$. 
For every other vertex this value is unknown, thus the event is initialized to have infinite cost. 
After all such events are inserted into the queue,
the minimal cost entry $(u,c,p)$ is removed 
 and if it is the goal, the process terminates.
If not, then for every neighboring edge $(u,v)$ that is collision free, 
the cost to reach $v$ via $u$ is computed.
If it represents a shorter path to reach $v$, than $v$'s current entry, $v$'s entry together with its location in the priority queue are updated.

%
%
%
%
%
%
%
%
%

\subsection{Properties of our cost function}

Recall that our cost 
does not have an
optimal substructure
(see Sec.~\ref{sec:introduction} and Fig.~\ref{fig:substructure}).
%
This implies that, for any vertex~$u$, we need to consider not only the optimal path to reach~$u$, but also all paths that pass through $u$ and may be part of a minimal-cost path to reach some future vertex $v$ (that pass through~$u$).
To do so, we must characterize this set of paths.
We start by noting the following properties of Eq.~\ref{eq:e-cost}:%
\begin{observation}
\label{obs:1}
Let $\gamma$ be a trajectory that lies completely within $\Csafe$, 
then $\forall t~\lambda_{\gamma}(t) = 0$ and
the cost of the trajectory is simply its duration $T_\gamma$.
\end{observation}
\begin{observation}
\label{obs:2}
Let $\gamma$ be a trajectory that lies completely within $\Crisk$, 
then $\forall t~\lambda_{\gamma}(t) = t$ and
the cost of the trajectory is $e^{T_\gamma}-1$.
\end{observation}

Using the fact that Obs.~\ref{obs:1} and~\ref{obs:2} hold for any maximally connected subpath in $\Csafe$ or $\Crisk$ we can rewrite Eq.~\ref{eq:e-cost}.
Namely, let 
$t_0 = 0 < t_1 < \ldots < t_n = T_\gamma$ such that $\forall i \geq 0$,
$\gamma[t_{2i}, t_{2i+1}] \subset \Csafe$ and
$\gamma[t_{2i+1}, t_{2i+2}] \subset \Crisk$
then\footnote{There is a slight inaccuracy here as \Crisk is an open set and thus $\gamma[t_{2i+1}, t_{2i+2}]$ cannot be fully contained in~$\Crisk$. However this inaccuracy does not change our results and is intentionally used for ease of exposition.} 
\begin{equation}
\label{eq:cost2}
c (\gamma) 
=
\sum_i {\left(t_{2i+1} - t_{2i}\right)} + 
\sum_i {\left(e^{t_{2i+2} - t_{2i+1}} - 1 \right)}. 
\end{equation}

We characterize the set of paths that our algorithm will have to consider using the notion of \emph{domination}. 
Given two trajectories $\gamma_1, \gamma_2$ that start at $x_s$ and end in some vertex~$u$, we say that~$\gamma_1$ \emph{dominates}~$\gamma_2$ if 
it will always be more beneficial to use $\gamma_1$ and not~$\gamma_2$ to reach some future vertex $v$.
Namely, $\gamma_1$ dominates~$\gamma_2$ if 
$c(\gamma_1) \leq c(\gamma_2)$ and
$\lambda_{\gamma_1}(T_{\gamma_1}) \leq \lambda_{\gamma_2}(T_{\gamma_2})$.
The set of all trajectories $\Gamma_u$ that start at $x_s$ and end in $u$ where no trajectory dominates any other trajectory is said to be a \emph{useful} set of trajectories. 

Understanding path domination and the maximal size of a useful set of trajectories will be key in understanding our algorithms and bounding their running time. We note several properties of such trajectories:


\begin{lem}
\label{lem:dominating}
Let $u \in \Csafe$ with $\gamma_u$ the minimal-cost path to reach $u$ from $x_s$.
Then $\gamma_u$ dominates any other trajectory $\gamma_u'$ that reaches $u$.
\end{lem}
\vspace{-6mm}
\begin{proof}
Since $u \in \Csafe$ we have that
$\lambda_{\gamma_u}(T_{\gamma_u}) = \lambda_{\gamma_u'}(T_{\gamma_u'}) = 0$.
Furthermore, $\gamma_u$ is a minimal-cost path, thus
$c(\gamma_u) \leq c(\gamma_u')$.
Hence, by definition, $\gamma_u$ dominates $\gamma_u'$.
\end{proof}

\begin{lem}
\label{lem:dominating2}
Let $u \in \Crisk$ with $\gamma_u$ the minimal-cost path to reach $u$ from $x_s$.
Let $\phi(u)$ be the point on the boundary of~$\Csafe$ through which~$\gamma_u$ last entered $\Crisk$.
Then, $\gamma_u$ dominates any trajectory $\gamma_u'$ that reaches $u$ with $\phi(u)$ as the last point on the boundary of $\Csafe$ through which~$\gamma_u'$ last entered~$\Crisk$.
\end{lem}
\vspace{-6mm}
\begin{proof}
Let $t_1$ and $t_1'$ be the times that $\gamma_u$ and $\gamma_u'$ reach~$\phi(u)$, respectively.
If $c(\gamma_u[0,t_1]) > c(\gamma_u'[0,t_1'])$, by Eq.~\ref{eq:cost2}, we can replace subpath  $\gamma_u[0,t_1]$ with $\gamma_u'[0,t_1']$ in $\gamma_u$ and reduce its cost which contradicts $\gamma_u$  being a minimal-cost path.
The same  argument holds for subpaths 
$\gamma_u[t_1, T_{\gamma_u}]$ and 
$\gamma_u'[t_1', T_{\gamma_u'}]$.
Thus, we have that 
$\lambda_{\gamma_u}(T_{\gamma_u})
\leq
\lambda_{\gamma_u'}(T_{\gamma_u'})$
and 
$\gamma_u$ dominates~$\gamma_u'$. 
\end{proof}

\section{Minimal-cost path-finding algorithm}
\label{sec:alg}

In this section we address problem P2, 
namely how to efficiently compute a minimal-cost path between two vertices in a given roadmap.
As we have seen, in Dijkstra's algorithms (and its many variants), 
each vertex $u$ can have only one useful path which will dominate all other paths to reach~$u$.
Lemma~\ref{lem:dominating} and~\ref{lem:dominating2} imply that in our setting this holds for vertices in \Csafe but for vertices in \Crisk, the number of useful paths may be as large as the number of edges entering \Crisk.
This gives rise to two different algorithmic approaches:

The first approach (Sec.~\ref{subsec:naive}) is to directly use Eq.~\ref{eq:cost2}
by splitting the graph into subgraphs that are fully contained in~\Csafe and subgraphs that are fully contained in~\Crisk. 
We then pre-compute the graph distance between all points that lie on the border of \Csafe and \Crisk (we call these \emph{border points}) restricted to moving only in \Crisk. 
Using these distances allows us to define a new graph, which has an edge between every pair of border points with weights assigned using the precomputed distances.
We can then run any shortest-path algorithm on the new graph without having to consider the multiple useful paths of vertices in~\Crisk.
This is analogous to having our snails roam the original graph and considering a traversal of \Crisk as one discrete event.

The first algorithm (Sec.~\ref{subsec:naive}), which requires preprocessing the entire graph, can be seen as a warm-up for our efficient path-finding algorithm (Sec.~\ref{subsec:efficient}).
This algorithm essentially runs a Dijkstra-type search without any preprocessing. 
To do so, for vertices within \Crisk, it efficiently maintains all useful paths. 
Here we can envision our snails entering \Crisk  as in Dijkstra's algorithm. However, when one snail reaches a vertex already traversed by another snail, it only retracts into its shell if the other snail dominates it.

Both algorithms have to distinguish between \Csafe and~\Crisk. Thus, we start by defining the \emph{refined roadmap} (Sec.~\ref{subsec:refined}) and then continue to detail each algorithm.

\subsection{Refined roadmap}
\label{subsec:refined}
An edge $e$ is said to be a \emph{border edge}
if it straddles \Csafe and~\Crisk.
Namely, if
$e \cap \Csafe \neq \emptyset$ and 
$e \cap \Crisk \neq \emptyset$.
For simplicity of exposition, we assume that every border edge of $G$ intersects the boundary of $\Crisk$ exactly once. We denote this point $\phi(e)$ and call it a \emph{border point}.
Set 
$E_{\text{border}} \subseteq E$ to be the set of all border edges and
$V_{\text{border}} = \bigcup_{e \in E_{\text{border}}} \phi(e)$ to be all the border points.

Given a roadmap $G = (V,E)$, define the \emph{refined roadmap} $\tilde{G} = (\tilde{V}, \tilde{E})$ such that
$\tilde{V} = V \cup V_{\text{border}}$
and
$\tilde{E} = 
(E \setminus E_{\text{border}}) \bigcup 
\{ (u,\phi(e)), (\phi(e),v)~\vert~(u,v) \in E_{\text{border}} \}$.
Namely, the refined roadmap is the roadmap defined by adding all border points to the original set of vertices and subdividing border edges accordingly.

\subsection{Minimal-cost planning via precomputinon }
\label{subsec:naive}

To compute the shortest path in $G$, 
we start by constructing~$\tilde{G}$.
For each border point, we run Dijkstra's algorithm restricted to $\Crisk$.
This gives us a mapping 
$\calT: V_{\text{border}} \times V_{\text{border}} \rightarrow \R_{\geq 0}$ that denotes shortest \emph{distances}, or traversal times, of paths that stay strictly in~$\Crisk$ 
(if no such path exists, then the mapping returns $\infty$).
Thus, the cost of the shortest path between two border points $u,v$ that stays strictly in $\Crisk$ is $e^{\calT(u,v)}-1$.
Now, construct the graph $G' = (V', E')$ where
$V' = (V \cap \Csafe) \bigcup V_{\text{border}}$,
and
$E' = 	(E \cap \Csafe)
		\bigcup 
		\{ (u,\phi(e))~\vert~(u,v) \in E_{\text{border}} \}
		\bigcup 
		\{(u,v)~\vert~u,v \in V_{\text{border}} \text{ and } \calT(u,v) < \infty\} $.
Namely, $V'$ consists of all vertices that are risk free or are border points. 
$E'$ contains three types of edges:
(i)~all original edges that are risk free, (ii)~ edges from $\tilde{G}$ that start at a risk-free vertex and end at a border point, and (iii)~new edges connecting each pair of border points within the same risk zone.
 The weights of edges in $G'$ are simply the weights of the edges in $\tilde{G}$ if they are of the first two types or $e^{\calT(u,v)}-1$ if the edge $(u,v)$ is of the third type.
 See Fig.~\ref{fig:refined}.
 Finally, we run any shortest-path algorithm between $x_s$ and $x_t$ in $G'$.
%
%

This algorithm requires preprocessing the entire graph and computing distances between all pairs of border points. As we will see, this may incur unnecessary computations. We continue with a Dijkstra-type algorithm that computes paths in a just-in-time manner.

\subsection{Minimal-cost planning via incremental search}
\label{subsec:efficient}
Recall that Dijkstra's algorithm makes use of a priority queue~$\calQ$ with entries of the form $(u,c,p)$.
Our algorithm will also store entries in~$\calQ$  which we call \emph{Risk-Aware Shortest Paths} entries, or RASP entries.
Each such RASP entry represents a dominating path.
Following Lemma~\ref{lem:dominating},
for each vertex $u \in \Csafe$ we need one such entry.
Following Lemma~\ref{lem:dominating2}, 
for each vertex $u \in \Crisk$ we need at most one entry for each border point of $u$'s risk region.

Specifically, each entry 
$\tau=(u, c, t, \lambda, p, \phi)$ 
will represent a dominating path, or trajectory $\gamma_\tau$, to reach a vertex~$u[\tau] = u$ 
(implicitly defined by the parent pointer $p[\tau] = p$), 
its cost and duration 
(stored as $c[\tau] = c$ and $t[\tau] = t$, respectively),
its current exposure time at time $t[\tau]$
(stored as 
$\lambda[\tau] = \lambda$)
and the last border point ($\phi[\tau] = \phi$) that $\gamma_\tau$ passed through if $u \in \Crisk$ (NIL if $u \in \Csafe$)%
.


The algorithm (Alg.~\ref{alg_search}) starts by initializing RASP entries (lines~1-4) and inserting them into the min-priority queue~$\calQ$ (lines~5-6).  
Entries in~$\calQ$ are ordered according to their cost.
We iteratively pop the min-cost entry $\tau$ from~$\calQ$ and set $u = u[\tau]$ to be the vertex associated with the entry (line~8).
If it is the goal vertex, then a minimal-cost path has been found and the algorithm terminates (lines~9-10).
If not, we consider each of its neighbors $v$, and if the edge connecting the two is collision free we expand the trajectory $\gamma_\tau$ to $v$ (line 12). 
This trajectory $\gamma_{\tau_{\text{tmp}}}$ is represented by a temporary RASP entry~$\tau_{\text{tmp}}$.
As in Dijkstra's algorithm, we check if it improves the current-best path to reach $v$. Here we restrict ourselves to paths that enter $\Crisk$ through a specific border point $\phi[\tau_{\text{tmp}}]$.
If this is the case (line 13-15), we update the relevant RASP entry and decrease it's cost in~$\calQ$ (lines~14-15)

We now detail the \texttt{expand} operation (line~12).
Specifically, let $\tau$ be the entry popped from $\calQ$ with $u = \tau[u]$ its associated vertex. Let $v$ be its neighbor
and let $\Delta_t(e)$ denote the length of an edge $e$. 
We describe the content of the new RASP entry $\tau_v$ according to whether~$u$ and $v$ are in $\Csafe$ or~$\Crisk$.
See 
\arxiv{appendix}{extended version of this paper~\cite{SHS17}}
for a visualization of how the RASP lists are maintained by the algorithm.

\textbf{Case (i) $\bm{u \in} \Csafe$ and $\bm{v \in} \Csafe$:}
We set 
$\tau_v = (
	u,
	c[\tau] + \Delta_t(u,v), 
	t[\tau] + \Delta_t(u,v),
	0,
	\text{NIL},
	\text{NIL})		
$.

\textbf{Case (ii) $\bm{u \in} \Csafe$ and $\bm{v \in} \Crisk$:}
We compute the border point $\phi = \phi(u,v)$ and 
the lengths $\Delta_t(u, \phi)$ and $\Delta_t(\phi, v)$.
We then compute the RASP entry which represents the path reaching $v$ using $u$ as its parent.
This entry~$\tau_v$ will have,
$c[\tau_v] = c[\tau] + \Delta_t(u, \phi) + e^{\Delta_t(\phi,v)} - 1$,
$t[\tau_v] = t[\tau] + \Delta_t(u, v)$,
$\lambda[\tau_v] = \Delta_t(\phi,v)$ and
$\phi[\tau_v] = \phi$.

\textbf{Case (iii) $\bm{u \in} \Crisk$ and $\bm{v \in} \Crisk$:}
We set
$c[\tau_v] = c[\tau] + e^{\lambda[\tau]} \cdot (e^{\Delta_t(u,v)}-1)$,
$t[\tau_v] = t[\tau] + \Delta_t(u,v)$,
$\lambda[\tau_v] = \lambda[\tau] + \Delta_t(u,v)$ and
$\phi[\tau_v] = \phi[\tau]$.

\textbf{Case (iv) $\bm{u \in} \Crisk$ and $\bm{v \in} \Csafe$:}
We compute the border point $\phi(u,v)$ and 
the lengths $\Delta_t(u, \phi(u,v))$ and $\Delta_t(\phi(u,v), v)$.
Similar to case (ii) we set
$c[\tau_v] = c[\tau] + e^{\lambda[\tau]} \cdot (e^{\Delta_t(u,\phi(u,v))}-1) + \Delta_t(\phi(u,v),v)$,
$t[\tau_v] = t[\tau] + \Delta_t(u, v)$,
$\lambda[\tau_v] = 0$ and
$\phi[\tau_v] = \text{NIL}$.

%

\begin{algorithm}[tb]
\caption{\texttt{incremental\_search} ($G, x_s, x_g$)}
\label{alg_search}
\begin{algorithmic}[1]
 \STATE $\tau_{x_s, \text{NIL}} = (x_s, 0, 0,0, \text{NIL}, \text{NIL})$; \hspace{0.5mm}
 		$T \leftarrow \{ \tau_{x_s, \text{NIL}}\}$
 \FORALL {$v \in V \setminus \{x_s \}$}
 	\FORALL {$\phi \in V_\text{border}$}
 \STATE $\tau_{v, \phi} = (v, \infty, \infty,\infty, \text{NIL}, \phi)$; \hspace{0.5mm}
 		$T \leftarrow T \cup \{ \tau_{v, \phi}\}$
  	 \ENDFOR
 \ENDFOR

 \FORALL {$\tau \in T$}
 \STATE $\calQ$.\texttt{add\_with\_priority}$(\tau)$
 \ENDFOR
 
 \vspace{3mm}
 
 \WHILE {$|\calQ|>0$}

 	\STATE $\tau \leftarrow \calQ.\texttt{extract\_min}()$; \hspace{0.5mm}
 	$u \leftarrow u[\tau]$
	\IF {$u = x_g$}
		\RETURN {\texttt{extract\_path($\tau$)}}
	\ENDIF 	

 \vspace{3mm}

	\FORALL {$v \text{ s.t. } (u,v) \in E \text{ and } (u,v) \notin \Cforb$}
	 \STATE $\tau_{\text {tmp}} \leftarrow \texttt{expand}(\tau, v)$; \hspace{0.5mm}
	 $\tau_{v} \leftarrow \tau_{v, \phi[\tau_{\text {tmp}}]}$
 	 
	\IF {$c(\tau_{\text {tmp}}) < c[\tau_v]$} 
	 \STATE $\tau_v \leftarrow \tau_{\text {tmp}}$
	 \STATE $\calQ$.\texttt{decrease\_priority}$(\tau_v)$
	\ENDIF 	
	\ENDFOR

 \ENDWHILE
\end{algorithmic}
\end{algorithm}


\subsubsection{Practical implementation}
Similar to our descriptions of Dijkstra’s algorithm (Sec.~\ref{sec:prelim}), we traded practical efficiency with ease of exposition (this has no effect on the asymptotic runtime of the algorithm). 
In practice, we can initialize $\calQ$ to contain only the RASP entry associated with $x_s$.
Other RASP entries can be created on the fly only when they are first constructed by the \texttt{expand} operation (lines 12-15).

Additionally, 
Lemma~\ref{lem:dominating2} 
states
that a given vertex can have at most $n_B$ useful trajectories. In practice, this number may be much smaller. 
Thus, before inserting a RASP entry $\tau$ to~$\calQ$, we can check if it is dominated by any other entry $\tau'$ in $\calQ$ with $u[\tau] = u[\tau']$.

Finally, the same algorithm can be transformed into an A*-type algorithm by using a heuristic  that estimates the cost-to-go and ordering $\calQ$ according to the sum of the cost-to-come and the estimated cost-to-go.

\subsection{Computational complexity}
\label{sec:complexity}
In this section we discuss the computational complexity of our search algorithms. 
We assume that testing if an edge is collision free and computing border points and distances take constant time.
We note that while this assumption is common in search algorithms such as Dijkstra and A*, in many motion-planning applications, these operations often dominate the (practical) running time of search algorithms~\cite{L06}. 

Recall that $n_B \leq m$ is the number of border points in~$G$ and that our algorithm that uses precomputation (Sec.~\ref{subsec:naive}), 
runs 
(i)~Dijkstra's algorithm from every border point (restricted to vertices within within \Crisk)
(ii)~adds an edge between every two border points in the same connected component and
(iii) runs a shortest-path algorithm on the new graph $G'$.
Step (i) takes $O\left( n_B \cdot ((n + n_B) \log (n + n_B) + m) \right)$ time. 
We then add $O(n_B^2)$ edges to our new graph in step (ii).
This implies that the number of vertices $n'$ and edges $m'$ of $G'$ is $n' = O(n + n_B)$ and $m' = O(m + n_B^2)$.
Thus, step (iii) takes 
$O(n' \log n' + m')$.
To summarize, our precomputation-based algorithm takes
$$
O\left(
n_B n \cdot \log n
+
n_B^2 \cdot \log n
+
n_B m 
\right).$$

Our incremental search algorithm (Sec.~\ref{subsec:efficient}) has complexity identical to Dijkstra's except that there may be at most~$n \cdot n_B$ RASP entries in $\calQ$ (and not $n$). 
Moreover, each outgoing edge of a vertex $u$ can be expanded once for each of $u$'s useful paths which is at most~$O(n_B)$. 
Thus, our incremental search algorithm  runs in time,
$
O\left((n\cdot n_B) \log (n\cdot n_B) +  n_B\cdot  m\right)
$ which is equal to
$$
O\left(n_B n \cdot \log n +  n_B m\right).
$$

Interestingly, the (asymptotic) running time of the two algorithms is identical unless the $n_B^2 \log n$ component dominates the running time of the first algorithm.
This happens when $n_B \log n= \omega(m)$.
However, in practice, in the precomputation of the first algorithm we often compute paths in $\Crisk$ that will not be used. 
Moreover, this requires  testing in advance which edges are in $\Crisk$, which is an expensive operation in practice. This is demonstrated in 
Sec.~\ref{sec:eval}.

%

\section{Evaluation}
\label{sec:eval}

In this section we visualize our cost function and demonstrate the behavior of our algorithm.
All algorithms were implemented using the Open Motion Planning Library (OMPL 1.2.1)~\cite{SMK12} running on a 4.0-GHz Intel Core i7 processor with 16 GB of memory.
Source code is publicly available
at 
\url{https://github.com/personalrobotics/ompl_rasp}
.


For each experiment, we constructed a roadmap and precomputed for each vertex and each edge whether it is
 collision-free and whether it is in the risk zone.
This allows us to compare the time that graph operations take for each of the algorithms.


\begin{figure}[t]%
  \centering%
  \subfigure[]
  {
  \includegraphics[
  width=0.22\textwidth]{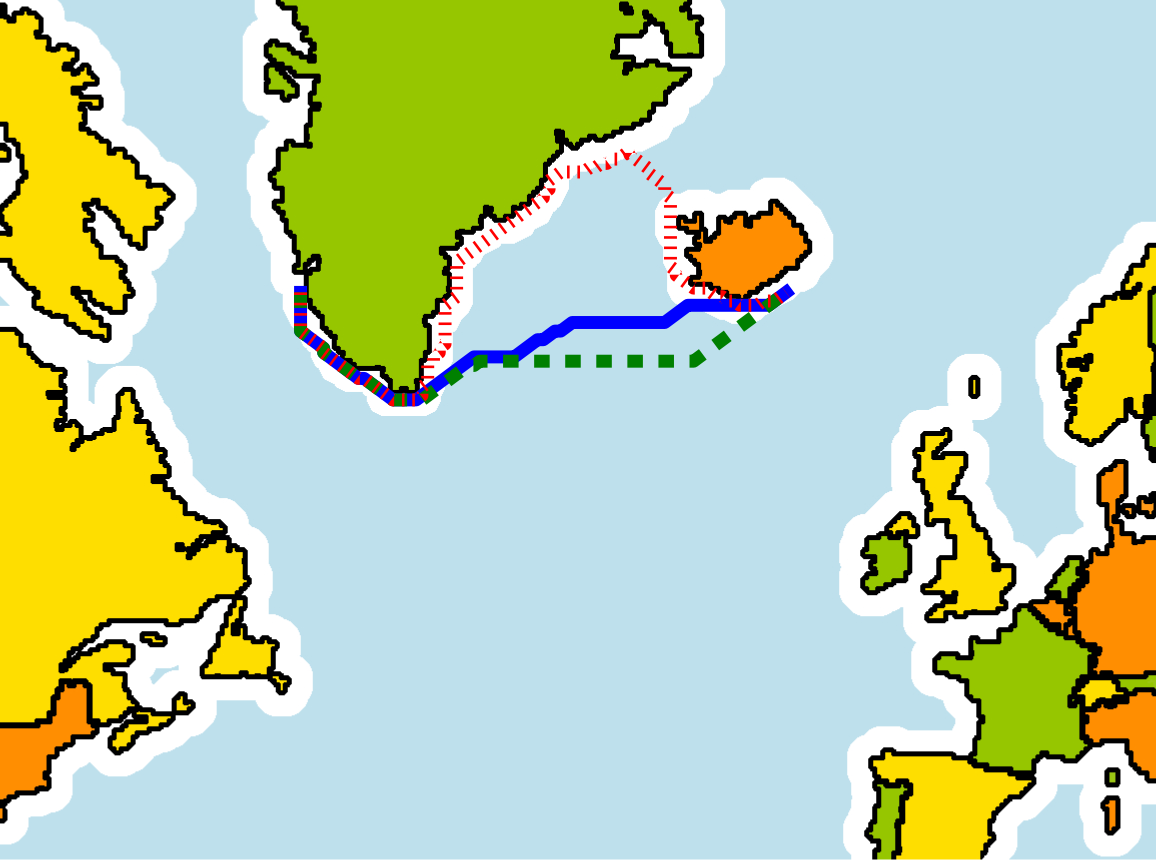}
  \label{fig:iceland}
  }
  \subfigure[]
  {
  \label{fig:norway}
  \includegraphics[
  width=0.22\textwidth]{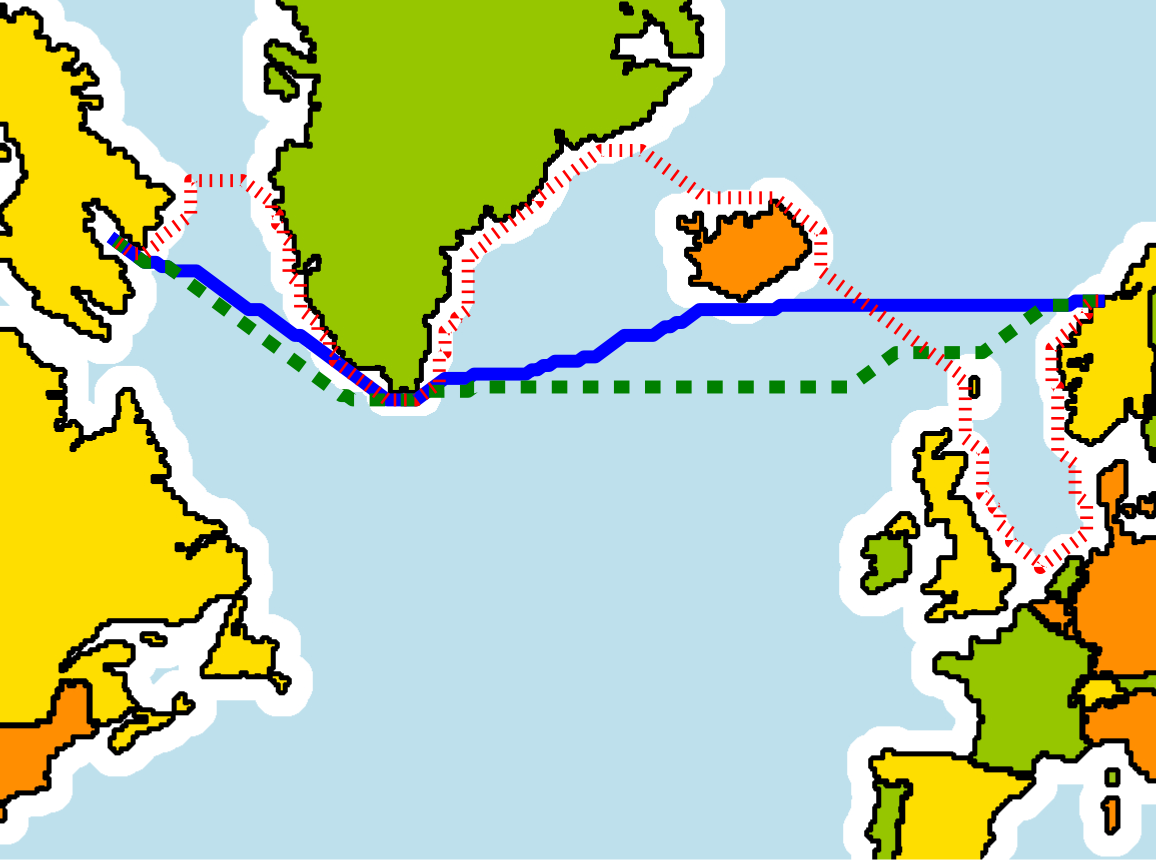}
  }
  \caption{%
    \captionstyle
    		Navigating the seas: land masses are regarded as obstacles, regions next to the coastal line (white) and open seas (light blue) are \Csafe and \Crisk, respectively.
    		We visualize paths produced using our cost function (solid blue), shortest paths (dashed green), and minimal-risk paths (dotted red).
    		Figure best viewed in color.
  }%
  \label{fig:Vikings}%
  \vspace{-5 mm}
\end{figure}

Our first set of experiments is motivated by the early Viking sailing expeditions: 
For centuries, sailing was done primarily by coastal navigation, where the sea vessel stayed within sight of the coast. 
Gradually, the art of open-seas navigation was developed, relying on more uncertain factors such as visibility to the sun, moon and the stars. 
Thus, we model the sea and the land as the free and forbidden regions, respectively.
Any point closer (further) than a predefined distance from the shore is modelled as the safe (risk) region, respectively.

Fig.~\ref{fig:Vikings} depicts maps with different queries.
For each query, 
we use a $201 \times 201$ eight-connected grid as a roadmap.
We then compute 
the minimal-cost path computed using Eq.~\ref{eq:e-cost},  
the shortest (Euclidean) paths and 
the minimal-risk path that minimizes time spent in~\Crisk.
As we can see, our cost function serves as a natural interpolation between the two opposing metrics.

We present running times in Table~\ref{tbl:res}.
Computing minimal-cost paths results in larger computation times when compared to computing shortest paths. 
For our incremental-based algorithm, this is roughly a 4$\times$ or 5$\times$ slowdown.
For our precomputation-based algorithm, this is slower by a factor of several thousands.
Not surprisingly, the lion's share of the algorithm's running time is dedicated to computing shortest paths between pairs of points on the boundary of~\Crisk.

\begin{table*}[t!]
\begin{center}
    \begin{tabular}{|l  | c | c | c |}
    \hline
     Scenario & Dijkstra & Precomputation-based & Incremental-based \\ \hline
     Vikings (Fig.~\ref{fig:iceland})
    		& $0.03 \pm 0.001$ 
    		& $204.35 \pm 8.73$ 
    		& $0.11 \pm 0.006$ \\ 
    	 Vikings~(Fig.~\ref{fig:norway})
    		& $0.06 \pm 0.004$ 
    		& $207.42 \pm 9.97$ 
    		& $0.31 \pm 0.034$ \\ 
     Assistive~Fig.~\ref{fig:example}
    		& $0.113 \pm 0.008$ 
   		& ---
   		& $0.003 \pm 0.008$  \\ 
\hline  		
    \end{tabular}
    \caption{
    \captionstyle
    Running time (in seconds) comparing Dijkstra, computing the  shortest path, with our precomputation-based algorithm (Sec.~\ref{subsec:naive}) and our incremental algorithm (Sec.~\ref{subsec:efficient}) computing minimal-cost paths. 
    For the precomputation-based algorithm, roughly $98 \%$ of the running time is spent on computing distances between pairs of boundary points on the Viking scenarios while on the assistive care scenario it terminated due to insufficient memory.
    Times reported are the average over 50 different runs together with one standard deviation.}
\label{tbl:res}
\vspace{-5mm}
\end{center}
\end{table*}

\begin{figure}[t!]%
  \centering%
  \subfigure[]
  {
  \label{fig:a}
  \includegraphics[
  trim={15cm 10cm 10cm 6cm},clip,
  width=0.2225\textwidth]{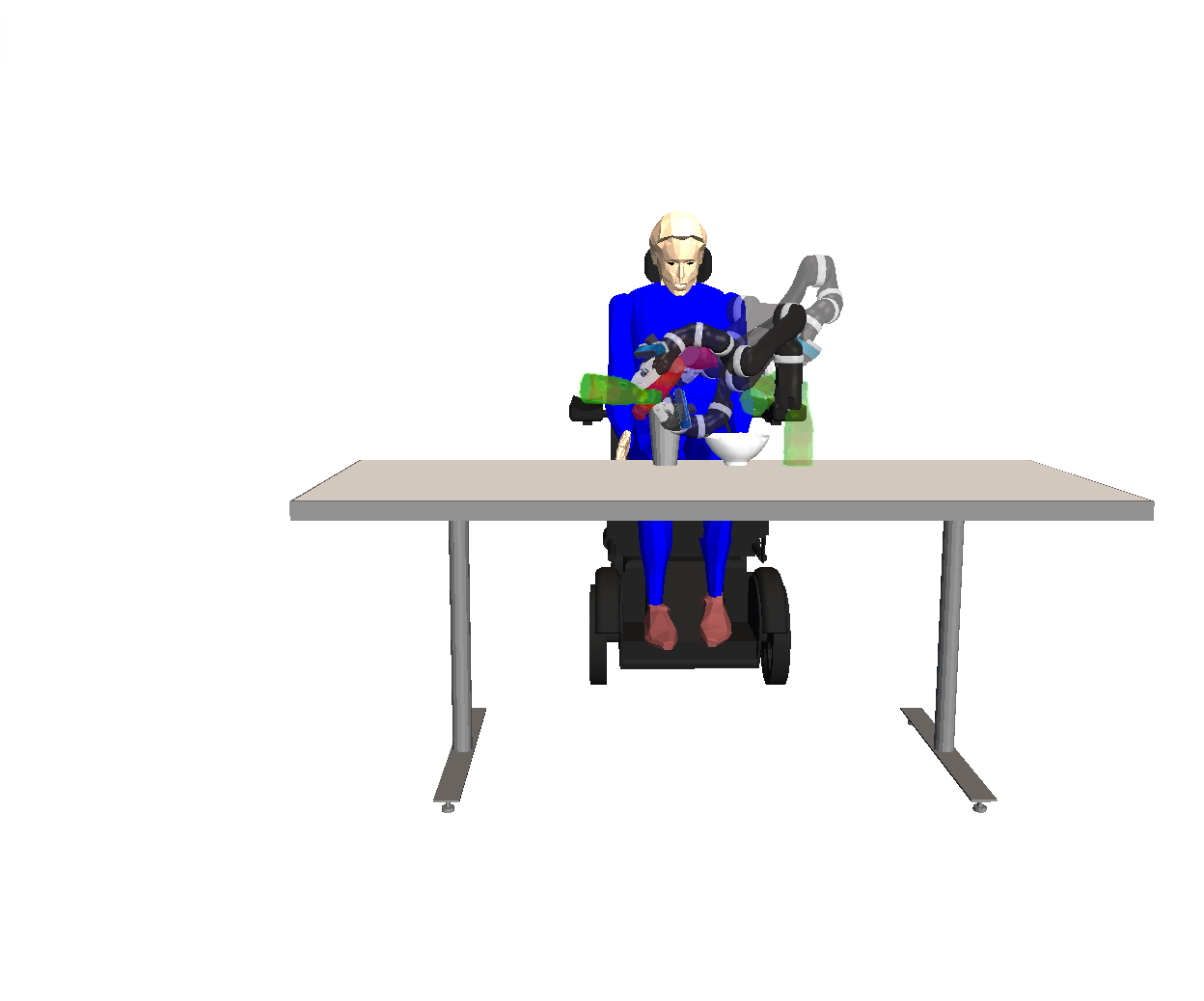}
  }
  \subfigure[]
  {
  \label{fig:b}
  \includegraphics[
    trim={15cm 10cm 10cm 6cm},clip,
    width=0.2225\textwidth]{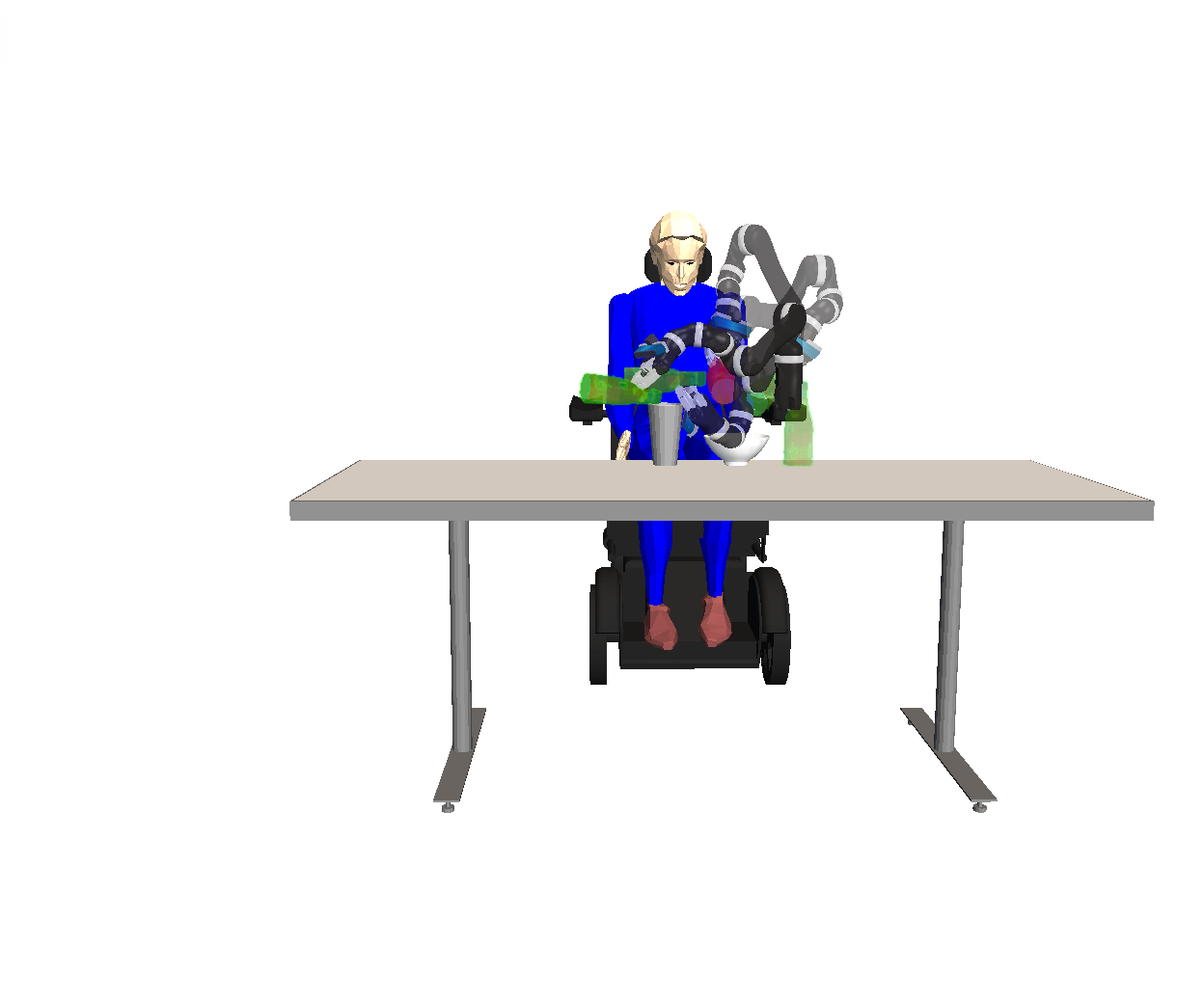}
  }
  \vspace{-5mm}
  \caption{%
    \captionstyle
       A disabled user moving a bottle using a robotic arm in the presence of obstacles. The trajectory of the arm moves between ``safe'' and ``risk''  regions where the bottle  is visible (colored green) and non-visible (colored red) to the user, respectively. 
Snapshots are taken at intermediate points along the path.
\subref{fig:a} Shortest path.
\subref{fig:b} Minimal-cost path.
  }%
  \label{fig:example}%
  \vspace{-5mm}
\end{figure}

Our second scenario is motivated by assistive robotics. Consider a robot arm performing a task such as pouring juice from a bottle, while receiving inputs from a user such as when to stop pouring. During the motion, the robot's end effector is moving between regions that are either visible or occluded to the user.

Specifically, in this motion-planning problem  configurations in \Crisk  are points occluded  from the viewpoint of a user sitting in a wheelchair.
We computed a Halton graph with 10,000 vertices 
(a typical-size roadmap in such motion-planning settings)
and 
ran Dijkstra's algorithm as well as our path-finding algorithms. Results, depicted in Fig.~\ref{fig:example} demonstrate how the shortest path traverses the risk zone for a long duration while minimal-cost paths enter it for a short period of time.

Timing results, reported in Table~\ref{tbl:res} show that our incremental-based algorithm is actually faster than Dijkstra's algorithm.
This is because in this scene there is a large risk region with high cost and a ``narrow passage'' that reduces risk exposure. Our cost function naturally guides the search towards this promising region. In contrast, Dijkstra searches in cost-to-come space and exposes more vertices.
Our precomputation-based algorithm, on the other hand, terminated due to insufficient memory.

\section{Conclusions and future work}
\label{sec:conclusions}

Many interesting research questions arise from our problem formulation. 
The first, relates to the roadmap generation: 
Using~\cite{KF11}, we can describe the necessary conditions for solutions obtained using PRM to converge to an optimal solution.
Applying the same analysis to our setting is not straightforward.
This is partially due to the fact that the proof used in~\cite{KF11} assumes that the source and target configurations are in the roadmap. 
Consider an optimal path $\gamma*$ that passes in and out of risk zones. 
A naive attempt to use the aforementioned proof is to subdivide the path into sections that are fully contained within \Csafe and fully contained within \Crisk and argue that asymptotically, the roadmap will converge to each of these subpaths.
However, the points where~$\gamma*$ moves from~\Csafe to ~\Crisk (and vice-versa) are \emph{not} in the roadmap.
We believe, that under certain assumptions on the structure of~\Crisk  this may be done but many details should be carefully addressed.


Since our problem is single-shot, a possible approach to solve RAMP (and not dRAMP) is not to construct a roadmap but a tree, rooted at the initial configuration. Here, an RRT*-type algorithm~\cite{KF11} may be used in order to asymptotically converge to the optimal solution.

Another interesting question relates to the  setting where $\calX \subset \R^2$,
i.e., when planning is restricted to the plane.
Here, we are interested in understanding what complexity class does our problem fall in.
It is well known that planning for shortest paths in the plane amid polygonal obstacles can be computed in $O(n \log n)$ time, where~$n$ is the  complexity of the obstacles (see~\cite{M04} for a survey).
When computing shortest paths amid 
polyhedral obstacles in $\R^3$,
or in $\R^2$ when there are constraints on the curvature of the path,
the problem becomes NP-Hard~\cite{CR87,KKP11}.
Furthermore, 
the Weighted Region Shortest Path Problem, 
which is closely related to our problem (Sec.~\ref{sec:related_work}),
is unsolvable in the Algebraic Computation Model over the Rational Numbers~\cite{DGMOS14}.
If our problem is NP-Hard, as we conjecture,  then a reduction, possibly along the lines of~\cite{CR87} should be provided together with an approximation algorithm.
Here, a possible approach would be to sample the boundary of \Crisk, similar to~\cite{AFS16}.

Finally, our work assumed that the dimension of \Csafe and~\Crisk are $d$, the dimension of $\calX$.
However, we envision our cost function being used in situations where this assumption does not hold.
Consider compliant motion planning or fine motion~\cite{LMT84} where a robot reduces uncertainty by making and maintaining contact with the environment. Here, we would like to penalize for \emph{not} being in contact with an obstacle feature.
Thus, \Csafe induces a manifold of lower dimensionality than $\calX$ which raises many interesting questions.

\arxiv{
\section*{Acknowledgements}
The authors thank Chris Dellin and Shushman Choudhury  for their contribution at the early
stages of the this work.
}{}

\arxiv{
\appendix
\section*{Appendix}
We visualize our incremental-based minimal-cost planning algorithm (Sec.~5.3) in Fig.~\ref{fig:rasp_filmstrip}.
Each sub-figure contains the current state of the priority queue~$\calQ$ (entries with unbounded cost are not shown) ordered from low cost (top) to high cost (bottom).
Additionally, each figure contains the current set of paths represented in the queue.
Paths are encoded (solid, dashed and dotted lines) according to the border point in their respective RASP entry.
For simplicity, we will use the terms paths and RASP entries interchangeably.

The algorithm starts  with~$\calQ$ containing one entry for the start vertex (Fig.~\ref{fig:rasp_a}).
It is then populated with two entries, representing paths to $x_1$ and $x_2$ (Fig.~\ref{fig:rasp_b}).
Since the path to $x_1$ has the minimal cost, it is popped from~$\calQ$ and the path to $y$ is added to~$\calQ$ (Fig.~\ref{fig:rasp_c}). Notice that this path, which traverses the risk region, is encoded using a dashed line to emphasize that the border point of the path's entry has changed.

Now, the path to $x_2$ is popped and an alternative path to reach $y$ is computed (Fig.~\ref{fig:rasp_d}).
Notice that now $\calQ$ contains two paths to reach $y$ but neither dominates the other. 
We pop the minimal-cost path to reach $y$ from $\calQ$ (dashed line from~$x_1$) and add to $\calQ$ the entry representing the path reaching~$z$ (Fig.~\ref{fig:rasp_e}).
Similarly, we pop the next-best path to reach $y$ from $\calQ$ (dotted line from $x_2$) and add to~$\calQ$ the entry representing the alternative path reaching~$z$ (Fig.~\ref{fig:rasp_f}).
Notice that now $\calQ$ contains two entries where the first dominates the second which, in turn, may be discarded.

\begin{figure*}[h!]%
  \centering%
  \subfigure[]
  {
  \includegraphics[width=0.3\textwidth]{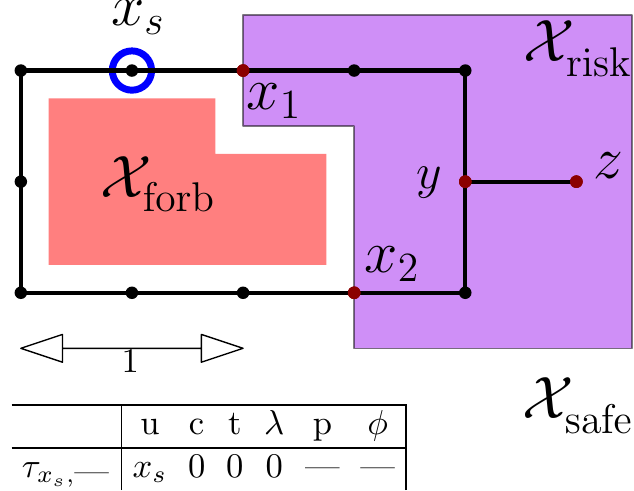}
  \label{fig:rasp_a}
  }
   \hspace{0mm}
  \subfigure[]
  {
  \label{fig:rasp_b}
  \includegraphics[width=0.3\textwidth]{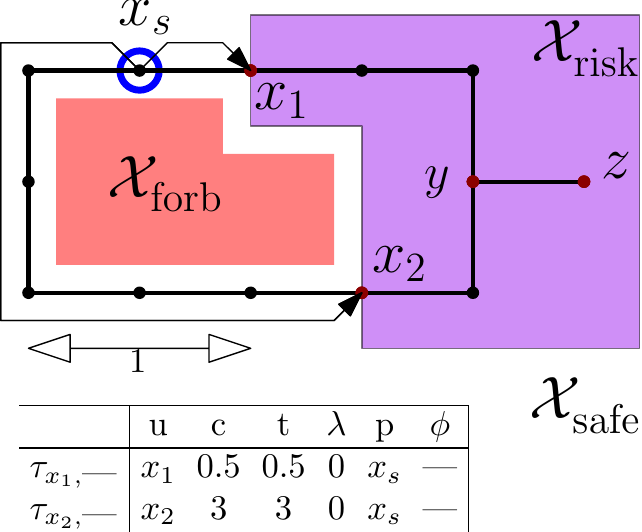}
  }
  \hspace{0mm}
  \subfigure[]
  {
  \label{fig:rasp_c}
  \includegraphics[width=0.3\textwidth]{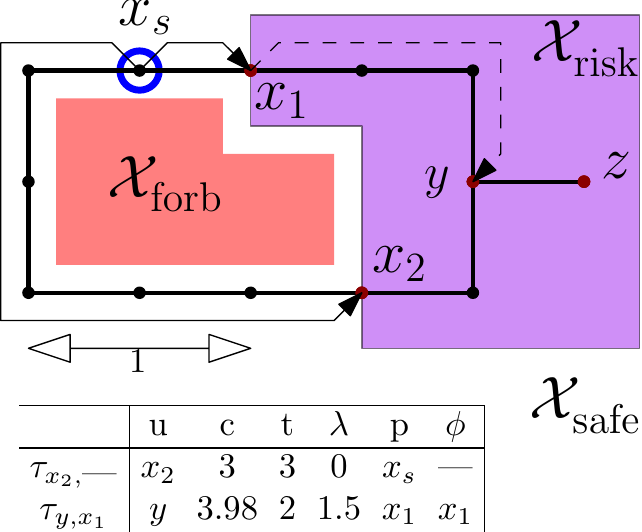}
  }  
  \hspace{0mm}
  \subfigure[]
  {
  \label{fig:rasp_d}
  \includegraphics[width=0.3\textwidth]{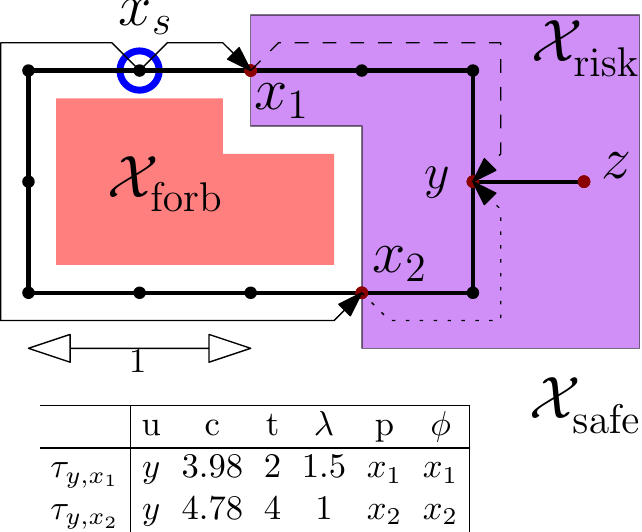}
  }
  \hspace{0mm}
  \subfigure[]
  {
  \label{fig:rasp_e}
  \includegraphics[width=0.3\textwidth]{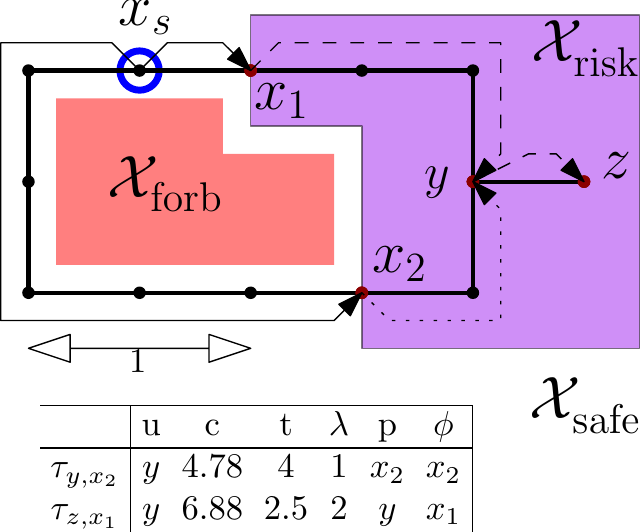}
  }
  \hspace{0mm}
  \subfigure[]
  {
  \label{fig:rasp_f}
  \includegraphics[width=0.3\textwidth]{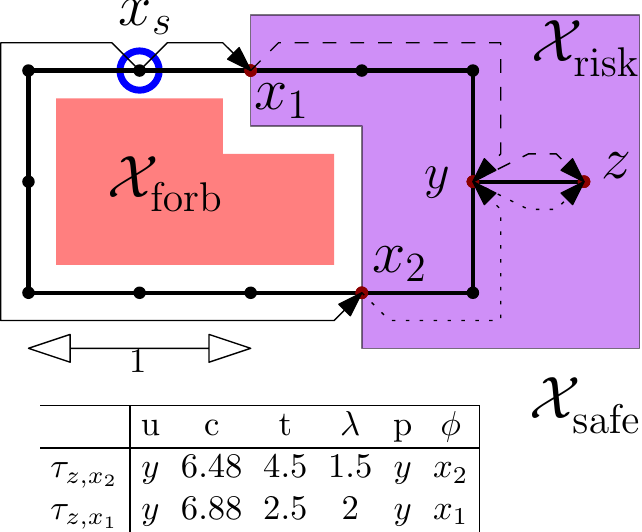}
  }    
  \caption{\captionstyle Visualization of incremental-based minimal-cost planning algorithm.}%
  \label{fig:rasp_filmstrip}%
\end{figure*}
}{} 


\end{document}